\documentclass{article}
\usepackage{arxiv}

\usepackage{geometry}
\usepackage{xcolor}
\usepackage{hyperref}
\usepackage{titlesec}
\usepackage{enumitem}
\usepackage{amsmath}
\usepackage{amssymb}
\usepackage{graphicx}
\usepackage{booktabs}
\usepackage{cite}
\usepackage{array}
\newcolumntype{C}[1]{>{\centering\arraybackslash}p{#1}}
\usepackage{xcolor}
\usepackage{colortbl}
\usepackage{comment}
\usepackage{subcaption}
\usepackage{makecell}
\usepackage{arydshln}
\usepackage{pifont}
\usepackage{xcolor}
\newcommand{\cmark}{\ding{51}}
\newcommand{\xmark}{\ding{55}}
\usepackage{booktabs}      
\usepackage{multirow}      
\usepackage{siunitx}       
\usepackage{makecell}      
\usepackage{amsmath}       
\usepackage{multirow}
\usepackage{siunitx}      
\definecolor{alrow}{RGB}{220, 230, 242}
\usepackage{pifont}
\definecolor{headerblue}{RGB}{220, 230, 242}
\definecolor{algreen}{RGB}{214, 234, 214}
\definecolor{improvegreen}{RGB}{0, 140, 0}
\definecolor{worsenred}{RGB}{180, 0, 0}
\definecolor{neutralgray}{RGB}{100, 100, 100}
\definecolor{bestrow}{RGB}{255, 243, 205}   

\definecolor{feedsblue}{RGB}{30, 80, 160}

\title{%
  \textbf{MUST-PET: MUltimodal Self-supervised learning across Tracers for whole-body PET/CT-based lesion segmentation}\\
}

\author{ 
	Bashirul Azam Biswas \\
Department of Biomedical Data Science\\
Geisel School of Medicine at Dartmouth\\
Hanover, NH 03755, USA\\
	\texttt{Bashirul.Azam.Biswas@dartmouth.edu} \\
    \And
    Amartya Bhattacharya \\
Department of Biomedical Data Science\\
Geisel School of Medicine at Dartmouth\\
Hanover, NH 03755, USA\\
	\texttt{Amartya.Bhattacharya.GR@dartmouth.edu} \\
	\And
	Biratal Raj Wagle \\
Department of Biomedical Data Science\\
Geisel School of Medicine at Dartmouth\\
Hanover, NH 03755, USA\\
	\texttt{Biratal.Raj.Wagle@dartmouth.edu} \\
    \And
	Matthew E. Maeder \\
Radiology\\
Dartmouth Hitchcock Medical Center\\
Lebanon, NH 03766 , USA\\
	\texttt{Matthew.E.Maeder@dartmouth.edu}\\ 
    \And
    James B. Yu \\
Radiation Oncology\\
Dartmouth Hitchcock Medical Center\\
Lebanon, NH 03766 , USA\\
	\texttt{James.B.Yu@dartmouth.edu}\\ 
    \And
	Indrani Bhattacharya \\
Department of Biomedical Data Science\\
Geisel School of Medicine at Dartmouth\\
Hanover, NH 03755, USA\\
	\texttt{Indrani.Bhattacharya@dartmouth.edu}\\
}
\date{}

\renewcommand{\shorttitle}{MUST-PET}

\begin{document}

\maketitle

\begin{abstract}
Deep learning–based automatic lesion segmentation in whole-body positron emission tomography/computed tomography (PET/CT) can help clinicians in staging, treatment planning, and response assessment. However, developing generalizable segmentation models remains challenging because of limited annotated data and domain shifts arising from differences in scanners, imaging protocols, institutions, and patient populations. Self-supervised learning (SSL) using large collections of unlabeled images has shown promise in addressing both label scarcity and domain generalization in medical imaging. Nevertheless, SSL remains underexplored in pan-cancer, multi-tracer, whole-body PET/CT, with existing approaches being limited to small and/or single-tracer datasets. In this work, we propose MUST-PET (MUltimodal Self-Supervised learning across Tracers), a multimodal, multi-tracer SSL framework for generalizable whole-body PET/CT lesion segmentation. MUST-PET is trained and validated with 5,910 pan-cancer, multi-tracer PET/CT scans from a diverse, multi-institutional dataset comprising public and institutional cohorts acquired with both $[^{18}\mathrm{F}]$ fluorodeoxyglucose ($[^{18}\mathrm{F}]$FDG) and prostate-specific membrane antigen (PSMA)-targeted radiotracers. MUST-PET leverages a SwinUNETR with a multimodal context-aware masked reconstruction objective, where both modalities are used as inputs, one modality (PET or CT) is randomly selected per batch, masked with zero-mean imputation, and reconstructed, with the other modality remaining fully visible. This pretraining objective allows multimodal context-aware reconstruction. The pretrained model is subsequently fine-tuned with labeled samples from the AutoPET III dataset. We evaluate MUST-PET via reconstruction quality assessment through mean absolute error, lesion segmentation performance, label-efficiency curves, and generalizability on three independent held-out test sets (AutoPET III-test with $N=321$, Deep-PSMA with $N=200$, and an institutional Dartmouth Hitchcock Medical Center dataset with $N=100$). MUST-PET reduces reconstruction error compared with the FDG-only baseline (for DHMC, Baseline vs. MUST-PET: 0.2970 vs. 0.2860) and substantially improves segmentation performance when compared with training from scratch (for AutoPET III, Baseline vs. MUST-PET: 0.526 vs. 0.576; for Deep-PSMA, Baseline vs. MUST-PET: 0.547 vs. 0.601). MUST-PET shows considerable performance gains with fine-tuning using limited labeled data and on unseen external datasets. These findings highlight the potential of multimodal, multi-tracer SSL to improve label efficiency and cross-domain generalization in whole-body PET/CT lesion segmentation.

\end{abstract}

\section{Introduction}
\label{sec:intro}

Automated pan-cancer, multi-tracer PET/CT-based lesion segmentation can assist clinicians in staging, treatment planning and response assessment, especially as PET/CT imaging volume grows exponentially without a corresponding increase in trained nuclear medicine specialists. 
However, accurate and generalizable models require large, labeled, multi-institutional datasets for supervised learning, and building such datasets is difficult, as annotation requires substantial clinician time and expertise.
Self-supervised pretraining on large unlabeled multi-tracer datasets could yield generalizable representations for downstream segmentation, but existing methods are similarly limited to single tracers and small datasets. Notable examples include Yazdani \textit{et al.}~\cite{yazdani2024automated} (SwinUNETR for PSMA-PET/CT lesion segmentation) and Patel \textit{et al.}~\cite{patel2022cross} (anomaly detection from healthy tissue). Self-supervised pretraining models like C²MAOT~\cite{huang2025c2maot} (cross-modal masked autoencoding for FDG PET/CT) and SegAnyPET~\cite{zhang2025seganypet} (a promptable model trained on 5,731 PET volumes) generalize well but are single-tracer or prompt-dependent, limiting full automation. More recent FDG-based foundation methods using masked autoencoding~\cite{oh2025fratmae,liu2026open} have not been evaluated for cross-tracer (e.g., PSMA), cross-disease, or unseen data generalizability. Drawing inspiration from the FDG-only foundation model~\cite{liu2026open}, we present a self-supervised pretraining method to learn robust pan-cancer, multi-tracer PET/CT features from a large, diverse, multi-institutional, multi-tracer, pan-cancer dataset.
Our proposed method, \textbf{MU}ltimodal \textbf{S}elf-supervised learning across \textbf{T}racers for whole-body PET/CT segmentation (\textbf{MUST-PET}), uses context-aware, multimodal masked encoding using unlabeled paired PET/CT images to learn robust feature representations, which are then fine-tuned using a whole-body lesion-segmentation model. We assess the representation quality of learned features through unlabeled multi-tracer reconstruction quality, data-efficient segmentation model refinement, and generalizability on unseen datasets and both FDG and PSMA tracers. Our method shows improved generalizability across tracers and datasets over the FDG-only model~\cite{liu2026open}. 


\section{Materials and Methods}
\subsection{Datasets}
\label{sec:datasets}

Our large and diverse dataset consists of $6{,}331$ PET/CT scans to train and evaluate our \textbf{MUST-PET} (Table \ref{tab:datasets}), consisting of four publicly available (AutoPET-III, DEEP-PSMA, SPADE, ViMED) and one internal Dartmouth Hitchcock Memorial Center (DHMC) datasets. AutoPET~III~\cite{gatidis2022fdg,ingrisch2024autopet} contains whole-body, pan-cancer, multi-tracer (FDG and PSMA) PET/CT scans with expert-annotated lesion segmentations. The Deep-PSMA dataset \cite{meakin_2025_17701815} consists of expert-annotated FDG and PSMA scans from metastatic prostate cancer patients acquired prior to LuPSMA therapy. The DHMC dataset includes de-identified, unlabeled PSMA-PET/CT scans from 1599 retrospective patients with or without prostate cancer who were seen at DHMC, Lebanon, NH. This retrospective study was approved by the Institutional Review Board (IRB) of Dartmouth College and Dartmouth Hitchcock Medical Centre, with patient consent waived. SPADE~\cite{eyuboglu2021weak} is a Stanford FDG dataset spanning routine staging and restaging cases, stored as pre-normalised
two-channel arrays. VI-MED~\cite{nguyen2025vietnamese} is a Vietnamese FDG dataset with demographic and technical variety from the North American and European sources. Its CT uses a non-standard integer range and its PET is stored as raw emission counts rather than SUV. 

Self-supervised pretraining uses the training and validation splits of AutoPET-III, DHMC, SPADE, and VI-MED, whereas segmentation fine-tuning uses the AutoPET-III training and validation sets. Pretraining is evaluated using reconstruction quality on the AutoPET-III and DHMC test sets in an unlabeled setting, while fine-tuning is evaluated on the labeled AutoPET-III and unseen Deep-PSMA test sets. As a preprocessing step, we resample CT and PET volumes to a common voxel spacing ($x=2, y=2, z=3$), crop to the body region and perform instance normalization, similar to \cite{liu2026open}.

\begin{table*}[htbp]
\small
\centering
\caption{Summary of datasets used. The FDG/PSMA row marks tracer availability (\cmark\ present, \xmark\ absent). Lu=Lung, Ly=Lymphoma, Me=Melanoma, Pr=Prostate, Neg=Negative.}
\label{tab:datasets}
\begin{tabular}{@{}lccccc@{}}
\hline
Properties & AutoPET-III & DeepPSMA & DHMC & SPADE & VI-MED \\

\hline
Disease dist.
 & {\begin{tabular}{c} Me/Lu/Ly/Pr/Neg \\ 188/168/145/537/573 \end{tabular}}
 & Pr  & Pr & -- & -- \\
\hline
FDG/PSMA              & \cmark/\cmark & \cmark/\cmark & \xmark/\cmark & \cmark/\xmark & \cmark/\xmark \\

\# lesions (per case) & 17.47$\pm$46.12 & 69.99$\pm$60.04 & 10.13$\pm$16.27 & -- & -- \\
\# slices (per vol)   & 135--963 & 195--1261 & 303--380 & 103-307 & 207-551 \\
Pretraining & Yes & No & Yes & Yes & Yes\\
Segmentation & Yes & Yes & No & No & No \\
Pretraining-train/val/test scans        & 1043/247/321 & 0/0/0 & 1350/149/100 & 1013/113/0 & 1417/578/0 \\
Segmentation-train/val/test        & 1043/247/321 & 0/0/200 & 0/0/0 & 0/0/0 & 0/0/0 \\
\hline
\end{tabular}
\end{table*}

\subsection{\textbf{MUST-PET}}

We employ SwinUNETR~\cite{swinunetr} for both self-supervised pretraining with unlabeled PET/CT imaging and supervised fine-tuning with labeled AutoPET-III training and validation cases. 

\subsubsection{Self Supervised Pre-training}

We extract two-channel PET/CT patches of size $96\times128\times128$ from each preprocessed PET/CT volume, yielding inputs  $\mathbf{X}\in\mathbb{R}^{2\times96\times128\times128}$. Each modality-specific patch is partitioned into non-overlapping masking blocks of size $12\times16\times16$. For each training sample, either the CT or PET modality is selected with equal probability, and 50\% of the blocks in the selected modality are randomly zero-masked while the complementary spatially aligned modality remains fully visible. The decoder reconstructs the masked voxels. This stochastic modality masking, together with cross-modal spatially aligned contextual information,  encourages the model to learn anatomical and functional representations, as well as relationships between the two modalities. The network is trained for 200 epochs using a weighted global reconstruction loss with the AdamW optimizer and cosine annealing learning schedular with a batch size of 3, an initial learning rate of $1\times10^{-4}$, and a weight decay of $1\times10^{-2}$. 

\subsubsection{Finetuning} 

SwinUNETR consists of (1) a Swin Transformer backbone, (2) convolutional encoder blocks, and (3) decoder blocks. Using the pretrained SwinUNETR described above, we investigate three supervised fine-tuning strategies for whole-body lesion segmentation: (1) full-model fine-tuning (\textbf{Full FT}), (2) encoder and decoder fine-tuning (\textbf{Enc. + Dec. FT}), and (3) decoder-only fine-tuning (\textbf{Dec. FT}). 
Following the nnUNet-based whole-body lesion segmentation on AutoPET-III dataset from Rokuss \textit{et al.}~\cite{rokuss2024fdg}, the segmentation models are optimized using an equally weighted combination of cross-entropy and Dice losses. To assess the effectiveness of pretraining under varying labeled data budgets, we finetune using 1\%, 5\%, 10\%, 20\%, 50\%, and 100\% of the available labeled training data. We evaluate all models on two held-out test sets (1) an in-distribution AutoPET-III-test set and (2) an out-of-distribution Deep-PSMA dataset.

\subsection{Evaluation Methods}
We evaluate SSL reconstruction quality by mean absolute error (MAE) between the preprocessed and reconstructed whole-body volumes, and evaluate lesion segmentation performance with Dice similarity coefficient, false-positive volume (FPVol), false-negative volume (FNVol), lesion-level sensitivity and predictive value (PPV)~\cite{wagle2026foundation}. 

\subsection{Experimental Design and Results}

\subsubsection{Comparing FDG-only pretraining with multi-tracer pretraining} We compare the reconstruction MAE of our model with that of the FDG-only pretrained model proposed by Liu \textit{et al.}~\cite{liu2026open}. The FDG-only model exhibits degraded reconstruction performance on PSMA scans (Table~\ref{tab:reconstruction_mae}), indicating limited cross-tracer generalizability. Another point that limits a fair comparison between the two models on the FDG scans of AutoPET-III is that we do not know the AutoPET-III split used in Liu \textit{et al.}~\cite{liu2026open}, and some of our AutoPET-III test scans could have been in their training set, biasing the model to the FDG-reconstruction. However, the PSMA cases in AutoPET-III and the DHMC cases were not used for training by any of the models, enabling fair comparison.

\begin{table}[!htbp]
    \centering
    \caption{Reconstruction MAE of the FDG-only~\cite{liu2026open} and our proposed MUST-PET model. Lower is better.}
    \label{tab:reconstruction_mae}
    \renewcommand{\arraystretch}{1.15}
    \setlength{\tabcolsep}{4pt}
    \footnotesize

    \begin{tabular}{@{}lcccc@{}}
        \toprule
        \multirow{2}{*}{\textbf{Model}}
        & \textbf{FDG}
        & \multicolumn{2}{c}{\textbf{PSMA}} \\
        \cmidrule(lr){2-2}
        \cmidrule(lr){3-4}
        
        & \textbf{AutoPET ($n=199$)}
        & \textbf{AutoPET ($n=122$)}
        & \textbf{DHMC ($n=100$)} \\
        \midrule
        FDG-only MAE
        & \textbf{0.2431}
        & 0.2982
        & 0.2970 
        \\
        
        MUST-PET (Ours)
        & 0.2709
        & \textbf{0.2908}
        & \textbf{0.2860} 
        \\
        \bottomrule
    \end{tabular}
\end{table}

\subsubsection{Label-efficient segmentation fine-tuning:} 
\begin{figure}[htbp]
  \centering
  \includegraphics[width=0.9\linewidth]{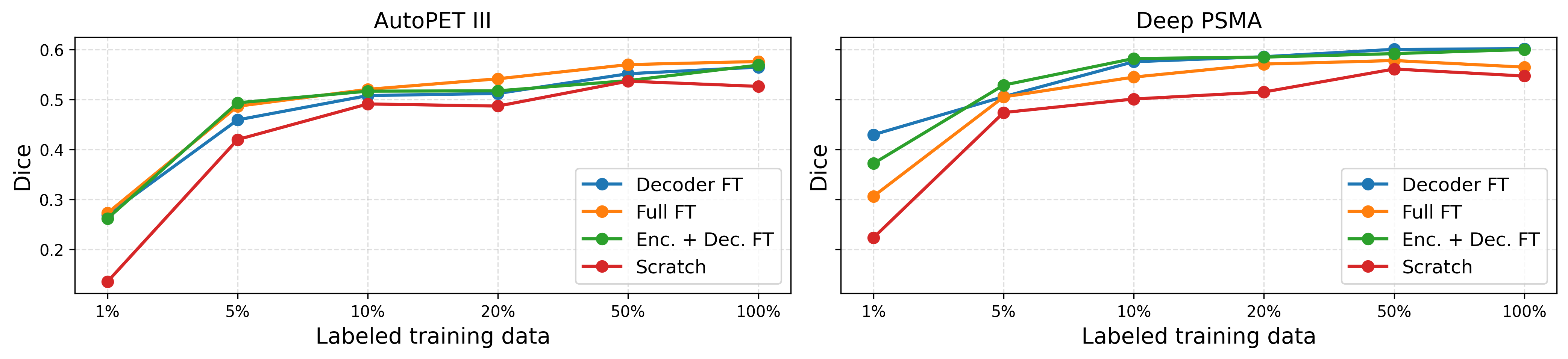}
  \caption{\textbf{Label-Efficient Fine-Tuning Performance.} Pretrained models provide substantial gains in low-label settings. Stronger generalisation is achieved in Deep PSMA dataset through the pretraining. 
}\label{fig:dice_data_efficiency}
\end{figure}

Self-supervised pretraining improves lesion detection across both datasets and all labeled-data budgets, with the largest gains observed under limited labeled-data settings (Figure~\ref{fig:dice_data_efficiency}), highlighting the utility of pretraining. 

\subsubsection{Lesion segmentation performance:} The \textbf{MUST-PET} models generally outperform scratch-trained SwinUNETR and nnUNet at both voxel and lesion levels (Table~\ref{tab:finetuning_comparison}). On AutoPET III, \textbf{Full FT} achieves the best Dice (0.576) and FP volume (14.112 mL), while \textbf{Decoder FT} provides the highest lesion sensitivity (0.794). On Deep-PSMA, \textbf{Decoder FT} achieves the best Dice (0.601), FN volume (5.684 mL), and lesion sensitivity (0.801), whereas \textbf{Enc.+Dec. FT} yields the lowest FP volume (36.397 mL).

\begin{table}[!htbp]
\centering
\caption{Lesion segmentation performance on the test sets with 100\% labeled AutoPET-III training data. FPVol and FNVol are reported in mL. $\uparrow$: higher is better; $\downarrow$: lower is better.}
\label{tab:finetuning_comparison}
\renewcommand{\arraystretch}{1.2}
\setlength{\tabcolsep}{6pt}

\begin{tabular}{c c c c c c c c}
\toprule
\textbf{Dataset}
& \makecell{\textbf{Model }}
& \makecell{\textbf{Training}\\\textbf{Strategy}}
& \textbf{Dice}
& \textbf{FP Vol}
& \textbf{FN Vol}
& \makecell{\textbf{Lesion}\\\textbf{Sensitivity}}
& \textbf{PPV} \\
& &
& ($\uparrow$)
& ($\downarrow$)
& ($\downarrow$)
& ($\uparrow$)
& ($\uparrow$) \\
\midrule


\multirow{4}{*}{AutoPET III}

    &
    SwinUNETR
    & Scratch
    & 0.526
    & 27.147
    & \textbf{7.586}
    & 0.773
    & 0.546 \\
    & nnUNet
    & Scratch
    & 0.566
    & 17.283 
    & 9.472
    & 0.773
    & \textbf{0.733} \\
    \cmidrule{2-8}
    & \multirow{3}{*}{\textbf{MUST-PET}}
    & Decoder FT
    & 0.565
    & 19.874
    & \underline{8.169}
    & \textbf{0.794}
    & 0.670 \\

    &
    & Enc.+Dec. FT
    & 0.569
    & 20.954
    & 8.457
    & 0.775
    & \underline{0.685} \\

    &
    & Full FT
    & \textbf{0.576} 
    & \textbf{14.112} 
    & 9.976
    & 0.778
    &  0.673 \\

    \midrule

    
\multirow{4}{*}{Deep PSMA} 
&
    SwinUNETR
    & Scratch
    & 0.547
    & 74.315
    & 9.785
    & 0.711
    & 0.732 \\
&    nnUNet
    & Scratch
    &0.561 
    & 73.030 
    & 7.933
    & 0.757 
    & \textbf{0.768} \\
    \cmidrule{2-8}
    & \multirow{3}{*}{\textbf{MUST-PET}}
    & Decoder FT
    & \textbf{0.601} 
    & 47.959 
    & \textbf{5.684}
    & \textbf{0.801}
    & 0.715 \\

    &
    & Enc.+Dec. FT
    & 0.600 
    & \textbf{36.397} 
    & 6.714
    & 0.779
    & 0.736 \\

    &
    & Full FT
    & 0.565 
    & 60.386 
    & 8.230
    & 0.751 
    & \underline{0.746}\\

    \bottomrule
\end{tabular}

\end{table}

 
 \begin{figure}[t!]
    \centering
\includegraphics[width=1\linewidth]{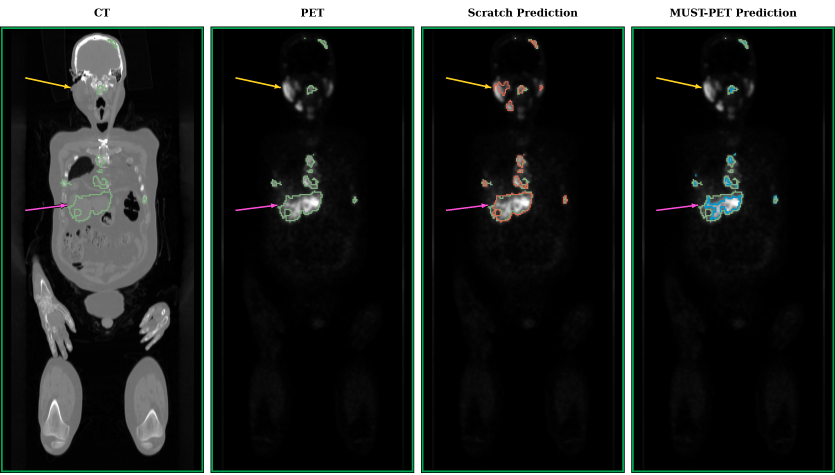}
\vspace{2pt}
    \caption{Qualitative comparison of lesion segmentation by the scratch and MUST-PET models, both trained on 100\% of the data. Green contours indicate GT, while red and blue contours show scratch and fine-tuned predictions, respectively. Yellow arrows indicate false positives from the scratch model, and pink arrows highlight improved lesion delineation with MUST-PET.}\label{fig:ct_pet_comparison}
\end{figure}






\subsubsection{Qualitative evaluation:}
\textbf{MUST-PET} reduces false positives and false negatives, while better capturing lesions, compared to training from scratch (Figure~\ref{fig:ct_pet_comparison}).

\section{Discussion and Conclusion}
In this work, we develop and test MUST-PET, the first context-aware, multimodal, multi-tracer self-supervised pretraining framework for whole-body PET/CT lesion segmentation using a large multi-institutional, pan-cancer dataset. MUST-PET achieves superior 3D whole-body reconstruction performance across both FDG and PSMA tracers, when compared to FDG-only foundation models, outperforms lesion segmentation from scratch, performs significantly better in lesion segmentation under limited labeled data budgets, and is 
generalizable across tracers, institutions, and different cancer types. 
A direct lesion-segmentation comparison between \textbf{MUST-PET} and the FDG-only model~\cite{liu2026open} was not feasible because the FDG-only model's training data overlap with our evaluation sets. Future work will include more extensive analysis of MUST-PET in other downstream tasks. 
\section*{Acknowledgements}
Research reported in this publication was supported by an Institutional Development Award (IDeA) from the National Institute of General Medical Sciences of the National Institutes of Health under grant number 1P30GM149408. We also gratefully acknowledge support from the Munck-Pfefferkorn Fund, the American Cancer Society Institutional Research Grant, the Department of Biomedical Data Science at Dartmouth College and the Department of Radiology at Dartmouth Hitchcock Medical Center. All of this support was instrumental in making this work possible.
Help from Claude Opus 4.8 was taken for correcting grammatical errors.

\bibliography{report} 
\bibliographystyle{spiebib} 

\end{document}